\documentclass[sigconf,anonymous=false,balance=true,review=false]{acmart}
\usepackage{booktabs} 
\usepackage{multirow}
\usepackage{verbatim} 
\usepackage{url} 
\usepackage{xcolor} 
\usepackage[T1]{fontenc}
\usepackage{lmodern} 
\usepackage{balance}
\usepackage{natbib}

\usepackage[inline]{enumitem}

\copyrightyear{2020}
\acmYear{2020}
\setcopyright{acmcopyright}
\setcopyright{rightsretained}
\acmConference[SAICSIT '20]{Conference of the South African Institute of Computer Scientists and Information Technologists 2020}{September 14--16, 2020}{Cape Town, South Africa}
\acmBooktitle{Conference of the South African Institute of Computer Scientists and Information Technologists 2020 (SAICSIT '20), September 14--16, 2020, Cape Town, South Africa}
\acmPrice{15.00}
\acmDOI{10.1145/3410886.3410899}
\acmISBN{978-1-4503-8847-4/20/09}

\begin{document}
\title{Complex Sequential Data Analysis: A Systematic Literature Review of Existing Algorithms}


\author{Kudakwashe Dandajena, Isabella M. Venter, Mehrdad Ghaziasgar and Reg Dodds}
\affiliation{
  \institution{Department of Computer Science, University of the Western Cape}
  \postcode{7535}
  \city{Bellville} 
  \state{Western Cape} 
}
\email{kudadandaz@gmail.com,{iventer,mghaziasgar,rdodds}@uwc.ac.za}

\renewcommand{\shortauthors}{Dandajena et al.}
\renewcommand{\shorttitle}{Complex sequential analysis with deep learning: a literature review}
\newcommand\rdcomment[1]{\textcolor{red}{---#1---\newline\noindent}}
\newcommand\rgcomment[1]{\textcolor{green}{---#1---\newline\noindent}}
\begin{abstract}
This paper provides a review of past approaches to the use of deep-learning
frameworks for the analysis of discrete irregular-patterned complex
sequential datasets.  A typical example of such a dataset is financial data
where specific events trigger sudden irregular changes in the sequence of
the data.  Traditional deep-learning methods perform poorly or even fail
when trying to analyse these datasets.  The results of a systematic literature review
reveal the dominance of frameworks based on recurrent neural networks.  The
performance of deep-learning frameworks was found to be evaluated mainly using
mean absolute error and root mean square error accuracy metrics.  Underlying
challenges that were identified are:  lack of performance robustness, non-transparency
of the methodology,  internal and external architectural design and configuration
issues.  These challenges provide an opportunity to improve
the framework for complex irregular-patterned sequential datasets.

\end{abstract}

%
%
%



\maketitle

\section{Introduction}

The art of improving the performance of any deep-learning framework is a process
of iterated refinements.  Currently,  there is no single ideal framework
that addresses the discontinuous, impulsive and irregular patterns of behaviour
associated with irregular-patterned complex sequential datasets~\cite{MaX19, Zha17}.
These extreme datasets can be found in many different domains, including: health
care, traffic, finance, such as stock prices, meteorology, such
as rainfall data and so forth. 
An automated artefact, beyond the conventional, suitable for solving prediction and
regression problems for such datasets will be useful for engineers and academics~\cite{Tan19}.
  
This paper considers the advances made towards the design and application
of frameworks for irregular-patterned complex sequential analysis based
on recent scholarly work.  

It was found that there are many deep-learning frameworks aimed at the
improvement of the analysis of sequential datasets but that there is no single
ideal framework for the analysis of irregular-patterned complex sequential 
datasets and that the frameworks that were developed were not extensively
evaluated using multidimensional evaluation mechanisms~\cite{Hew19}. Frameworks based
on recurrent neural network (RNN) architecture centred on long short-term memory
(LSTM)~\cite{Qin19} have been widely identified as the most suitable approach
towards addressing unstable sequential behaviour~\cite{Tan19}. 
It is empirically clear that the architectural designs of most present state-of
the-art sequential frameworks are simple extensions of the original LSTM
architecture~\cite{hur17}.  Most of these are equipped with gating mechanisms
for solving vanishing gradient problems.

An exhaustive literature search of all deep-learning related materials is not 
possible.  The influence of the foundational work by Hochreiter and Schmidhuber
in 1997 on the original LSTM architecture can be seen in
many of these recent research studies~\cite{Hoc97}.  Sequential forecasting
competitions have also contributed to the development of present state-of-the-art
benchmarks, methodologies, theories and datasets~\cite{Makr17}.
Some of the most influential competitions include the
M1 to M4 Competitions by Makridakis and Hibon, the Sante Fe competitions
by the Santa Fe Institute, the knowledge discovery and data mining (KDD) cup
competitions by the Association for Computing Machinery's Special Interest
Group on KDD, the Kaggle time series competitions by Goldbloom,
the global energy forecasting competitions by Tao Hong and the International
Journal of Forecasting~\cite{Hyn20}. 

Algorithm performance evaluation plays a critical role in the design 
of improved frameworks.  Many performance evaluation criteria or
mechanisms have been suggested.  These include: efficiency, accuracy,
consistency, reliability, stability, explainability, baseline 
comparison~\cite{Sok20}.  Most sequential deep-learning frameworks are 
evaluated using only accuracy as a criterion based on root mean square 
error as a metric~\cite{Meh16}.

It seems as if design and configuration issues of internal-, external- 
and hyper-parameters hinder the optimal performance of existing frameworks
for extreme datasets.  This points to the need for the development
of newer and more transparent frameworks that reveal
performance robustness in these environments.  
It is important to appreciate performance strengths and weaknesses of
existing novel frameworks on known datasets, before suggesting any new design
or architectural improvements to existing frameworks~\cite{Qia17}.

Using a systematic literature review, recent research publications were 
identified based on the following specific inclusion criteria: keywords, 
publication timelines, algorithms or framework relevance in terms of complex 
datasets, accreditation and citation quality of the journal.\\*[+6pt]
\noindent \textit{The survey explored the following questions}:\\
(1) How efficient are existing deep-learning algorithms for 
analysing complex sequential datasets?\\
(2) How should current deep-learning algorithms be adapted to deal with irregular-pat\-terned
    complex sequential datasets?\\

\noindent \textit{The primary aims of this systematic literature survey were to}:\\
(1) Identify well-known state-of-the-art deep-learning frameworks for complex sequential analysis.
\\
(2) Identify the challenges in current methods of analysing irregular-patterned
    complex sequential datasets.

This paper highlights the current state of affairs of deep-learning
frameworks for the transparent analysis of irregular-patterned complex 
sequential datasets and their challenges. Transparency in this instance 
refers to explainable frame-works or models that 
can be easily and independently replicated at any given time through 
experiments.

\section{Materials and methods}

\begin{figure}[!t]
\centering
\includegraphics[bb=16 24 568 525,scale=0.440]{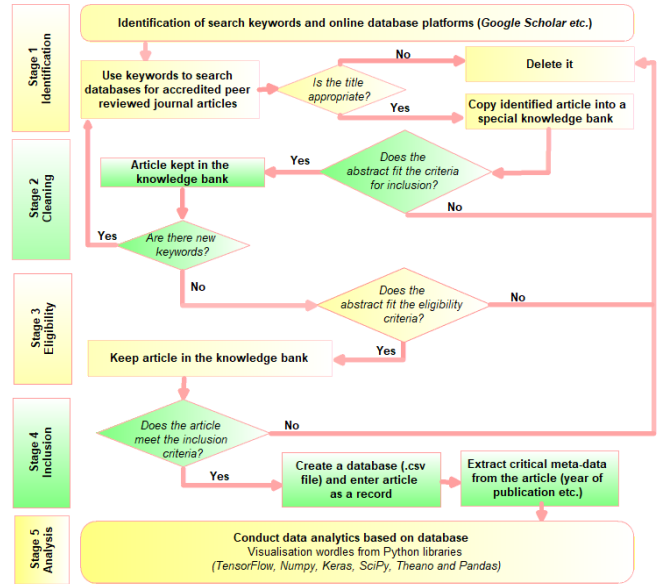}
\caption{Systematic literature review methodology}\label{figure:one}
\end{figure}

A systematic literature review, shown in Figure~\ref{figure:one}, was the
preferred methodology to demonstrate the breadth and depth of the existing
body of knowledge of deep-learning frameworks but also to identify inconsistencies
and gaps in this body of knowledge~\cite{Xia19}. 

According to Peter et al.~\cite{Pet19}, applying a systematic
literature review strategy offers a comprehensive, focused,
reliable, repeatable and thorough literature overview~\cite{And17,Laa17, Pic15}.

Referring to Figure~\ref{figure:one}, the systematic literature review 
methodology used in this research had the following phases:

\textit{Stage 1}---Identification of specific keywords to search for
irregular-patterned complex sequential analysis, as well as identifying appropriate
online research database platforms, for example Google Scholar. 
The identified keywords were used to search for published accredited peer reviewed
journal articles on online research database platforms~\cite{Pic15}.  The
title of the journal article was used to identify which articles to consider
and these were then stored for later evaluation. 

\textit{Stage 2}---Cleaning of the comprehensive list of the identified
journal articles that were collected during Stage 1.  This stage entailed
a qualitative evaluation of the fitness of each article based on the abstract
of the article.  The articles that did not pass the criteria for inclusion
were then deleted.  If new keywords were found in the abstracts, these were
added to the initial list of keywords and Stage 1 was repeated for these
new keywords,  in a \textit{grounded theory fashion}~\cite{Str97}.

\textit{Stage 3}---Eligibility of the articles was identified in Stage
2. A detailed qualitative screening exercise was applied to the output of
Stage 2, based on specific eligibility criteria~\cite{Sok201}. This stage
produced a comprehensive list of articles---those that did not adhere to
the eligibility criteria were discarded.

\textit{Stage 4}---Inclusion of the state-of-the-art articles was based
on inclusion criteria.  Articles identified in this stage were recorded 
in a database. For this stage the whole article was considered.
Critical meta-data was qualitatively identified and recorded in fields, for
example the year of publication, each with an appropriate column header.
The articles not included in this database were now discarded. The output
of this stage was a {\bfseries\texttt{.csv}} file.

\textit{Stage 5}---Analysis of the database using a quantitative method.
This empirical stage employed data analytic operations using Python libraries
such as TensorFlow, Numpy, Keras, SciPy, Theano and Pandas, within the Jupyter
Notebook environment.  This stage mapped the relationship between key features
of the records across identified fields.  During this stage the relationship
of records was visualised as \textit{wordles},  or word clouds. Finally,
the used codes, tools and the database of identified literature was uploaded
on an online platform to allow free access to other researchers.\footnote{All the 
experimental code is given in the Jupyter Notebook files on the GitHub website at:
\url{https://github.com/Dandajena/SDA/.}}

\begin{figure}[!t]
\centering
\includegraphics[bb=20 10 620 420,scale=0.470]{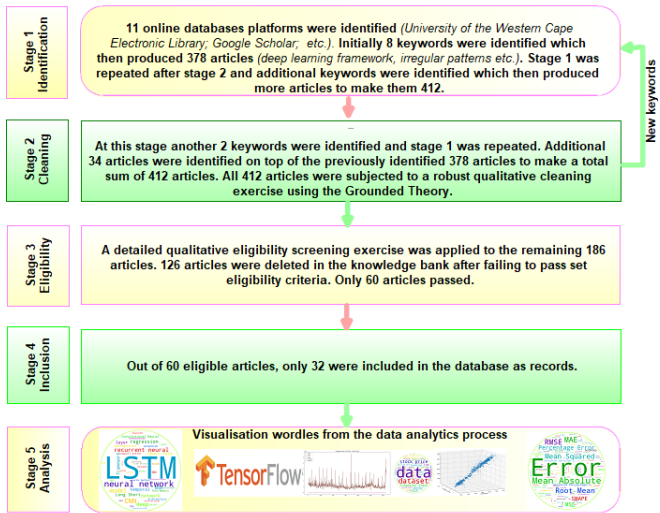} 
\caption{Processing the results of the systematic review}\label{figure:two}
\end{figure}

\section{Results}

The results are segmented into two sections. The first section highlights
the methodological implementation results which feeds into our last main
results section which provides the overall research findings of the systematic
literature review. 

\subsection{Methodological implementation results}

Figure 2 summarises the phases of the systematic literature review 
to provide the results of the implemented methodology.

\textit{Stage 1}---Initially eight search keywords were identified.  These
were: deep-learning framework, sequential algorithm optimisation, irregular
patterns, time series forecasting, parameter, volatile financial prediction
and extreme weather forecasting.  After the cleaning process of Stage 2,
two more keywords were identified, namely sequential learning and financial
signal processing.  Thus 10 keywords in total were used for identifying articles. 
The following 11 online research database platforms were used for the search:
\begin{anonsuppress}
University of the Western Cape 
\end{anonsuppress}
Electronic Library, Google Scholar, CiteSeerx,
GetCITED,  Microsoft Academic Research, Bioline International Directory of
Open Access Journals, PLOS ONE, Papers with Codes, BioOne, Science and Technology
of Advanced Materials, New Journal of Physics, ScienceDirect and NIPS. This
stage produced a total of 412 search results in the form of a comprehensive list
of peer reviewed articles. We focused on the results not exceeding the first
10 pages of each online database platform. These articles were stored for
further processing.

\textit{Stage 2}---Abstracts of 378 articles identified in Stage 1 were
subjected to a robust qualitative cleaning exercise using the \textit{grounded theory
approach}~\cite{Str97} which revealed two further keywords that were put through
Stage 1 again to identify more possible articles. This produced 34 more articles. 
Eventually 226 of the 412 articles were found not to be the right fit for
the study and were removed---186 articles were kept.

\textit{Stage 3}---The remaining 186 articles were evaluated based on eligibility
criteria: identified keywords, publication timelines from 2016 to date, algorithms
or framework relevance in terms of complex datasets, accreditation and quality
of the journal, i.e., its citation index~\cite{Sok20}. Abstracts of each
article were read in detail. A further 126 articles were trimmed off 
after failing to meet the eligibility standards---60 articles were kept.

\textit{Stage 4}---All 60 articles in the folder were now scrutinised and read in
depth. An excel \texttt{\bfseries.xlsx} database was created to capture a record of each article
that was deemed to be eligible.\footnote{Accessible at:
\url{https://github.com/Dandajena/SDA/blob/master/Database.xlsx.}}
Critical meta-data from the article were recorded in identified fields such
as:  journal source web link, journal name, journal title, authors, pages,
timelines (day, month, and year), editor, volume, issue number, city, country,
continent, standard number, day accessed, month accessed, year accessed,
data set, data set type, data set sources, dataset description, research
problem, research objective, implementation framework, architecture properties,
baseline models, best models, methodology, evaluation mechanism or criteria,
evaluation metric, results, novelty, future recommendations and gaps comment. 
Tabulated data fields were critically identified to suit empirical data analytics
operations for visualisation which would be implemented in Stage 5. 
A total of 28 ineligible articles were deleted in Stage 4 of the process since 
they were not a direct fit to the aim of the literature review study. Only 32 
articles were recorded in the database and the rest were deleted. The \texttt{\bfseries.xlsx} 
database file was further cleaned and converted to a
 \texttt{\bfseries.csv} file.\footnote{\raggedright Accessible at:
\url{https://github.com/Dandajena/SDA/blob/master/Database.csv.}}

\textit{Stage 5}---A program was implemented to analyse the database---the cleaned 
\texttt{\bfseries.csv} result of Stage 4.  The output of this program 
was the visualisation of the analysis in the form of graphs, wordles and 
schematic diagrams.  Finally, the code, tools and the literature
search database were uploaded to GitHub to allow open access of all 
experimental study material and code to other researchers. The literature 
research database can be freely accessed from the GitHub website.
Schematic diagrams were recorded as annexures.

\subsection{Systematic literature review results}

The most important finding from this study is the fact that current algorithms
cannot yet optimally analyse complex datasets. In most of the recent state-of-the-art
papers, Chinese publications dominate deep learning framework research.
The novelty of existing deep-learning frameworks is largely centred on their
application mechanism. From these publications it appears that researchers
normally design an architecture and then uniquely apply it to a particular
problem together with well-known architectures, such as generic recurrent
neural networks and long short-term memory to determine which of the architectures
is the most optimal for the specific application. We next discuss the results in 
terms of identified challenges, frameworks, datasets and their evaluation.

\subsubsection{Complex sequential analysis challenges}

The analysis of complex sequential datasets---characterised with spontaneous
or volatile behaviour---is far from attaining a maturity status.  According
to the literature, most modern deep-learning frameworks cannot address many
of the following challenges:
\begin{itemize}[leftmargin=\parindent]
\item Complexity in modelling and capturing extremely long-term sequential
patterns using traditional deep-learning algorithms such as RNN~\cite{Hua18}.
These algorithms lack transparency and explainability within the implementation
of deep-learning models~\cite{Cha18}.

\item
The analysis of highly variable, noisy and volatile datasets (the problematic
aspects of sequential datasets) Ma et al. (2019) ~\cite{MaX19}, Zhang et al. (2018) ~\cite{Zha185}
leads to performance disadvantages such as consistency or inconsistency,
sensitivity to outliers, extreme values and computing inefficiency~\cite{Bal16,Cha19}.

\item Most state-of-the-art deep-learning frameworks~\cite{Das18} experience
efficiency performance problems when exposed to different sequential datasets~\cite{Uma17}.

\item Model reliability and generalisation, is a problem when predictive
neural frameworks are used. These associated problems are caused by the stochasticity
of stock features in financial stock price datasets~\cite{Fen19}.

\item Precision challenges when forecasting within financial environments.
 This is associated with extreme sequential financial market datasets~\cite{Qia17}.

\item Lack of accurate, reliable, and interpretable modern deep-learning
models for uncertainty estimation over continuous variables~\cite{Kul18}.

\item Lack of a comprehensive comparison analysis of existing deep-learning
models for sequential learning.  Recent research focused on one-step forecasting,
based on smaller datasets~\cite{Qin19}.

\item Accurate sequential forecasting is a challenge when using existing
uncertainty estimation models.  This is a problem when dealing with its probabilistic
formulation which is difficult to tune, scale and it adds exogenous variables,
i.e., other variables outside the existing variables~\cite{Zhu17}.

\item The simultaneous forecasting of the inflow and outflow of crowds in
regions of a city is complex because of spatial dependencies, temporal dependencies
and external influence factors~\cite{Zha17}.

\item The complexity of sequential stock price datasets which require extensive
analysis resources~\cite{Xin19}.

\item The deficiency of traditional models towards the capturing of complex nonlinear
or dynamic dependencies between time steps and between multiple time series~\cite{Hua19}.

\item Predictive performance challenges associated with existing models and
the lack of well-established and explainable literature for sequential predictive
machine learning methods~\cite{Vit19}.

\item Dealing with robust and accuracy challenges which are associated with
existing sequential forecasting~\cite{Oro18} 

\item The ever-growing requirement of computing power, time and resources
of sequential forecasting modelling.  This is particularly associated with
unstable extreme weather patterns~\cite{Cha20}.  

\item Performance deficiency of sequential neural network models~\cite{Hew19}
when forecasting in multivariate dataset environments~\cite{Che19}.
\end{itemize}
\subsubsection{Identified sequential deep-learning frameworks}

To address some of the aspects of these research challenges and gaps, various
researchers have applied different techniques and methods to find solutions.
 The resources needed to deploy deep-learning frameworks with good performance,
is a problem as well as poorly configured internal and external parameters
and hyperparameters for such analysis.  Existing research work focuses on
the need to further explore and advance existing deep-learning models.
 Some of these are:  long short-term memory, adversarial LSTM, 
CapsNets, LSTM-convolutional neural networks, 
gated recurrent unit, and attention, bidirectional and temporal convolutional neural networks~\cite{Das18}. 

There is a need to explicitly combine the aspect of sequential dataset complexity
as an optimisation technique in the design of more advanced deep-learning
predictive algorithms or models or frameworks.  These frameworks require
a multidimensional evaluation mechanism which considers existing complex
datasets~\cite{Son18}.

In terms of on-going research, most researchers are citing the need to develop
state-of-the-art optimised algorithms that address challenges associated
with complex sequential environments.  There is a wide range of sequential
analysis frameworks as illustrated in Annexure 1 which were designed to resolve
these challenges. 

Figure 2 points to the fact that deep-learning frameworks, based on the recurrent
neural network architecture, long short-term memory by Hochreiter and Schmidhuber~\cite{Hoc97}
dominate the field of sequential forecasting.  
These gated architectures address exploding and vanishing gradient
problems associated with neural networks. Gated architectures are made up of
input, forget and output gates or modules to provide them with sequential learning
capabilities which decide which critical information to keep or discard during
sequential modelling. 
In sequential modelling, Tang et al.~\cite{Tan19} indicated that LSTM networks
are excellent for capturing features with longer sequences or time span unlike
the gated recurrent unit of Cho et al.~\cite{Cho14}.

\begin{figure}[!t]
\centering
\includegraphics[bb=20 230 370 590,scale=0.660]{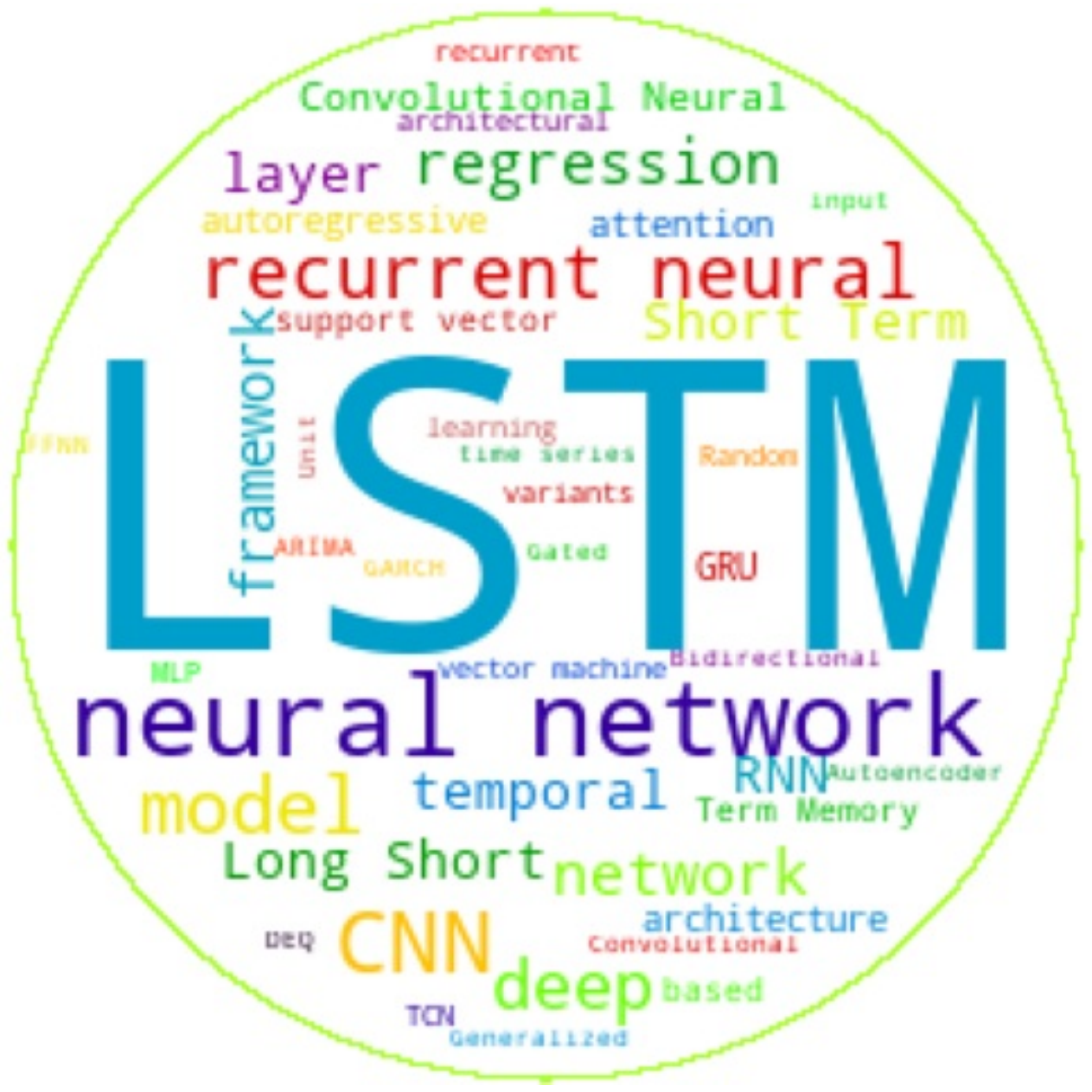}
\caption{Results of state-of-the-art deep-learning framework for sequential
challenges}\label{figure:three}
\end{figure}

According to Zhang et al.~\cite{Zha185} there are other competing modern
frameworks that can deal with this problem effectively.  These are: deep convolutional
neural networks (CNNs), capsule neural network, dilated RNN, dilated
CNN, ensemble CNN-LSTM-attention, DEQ-trellisNet and DEQ-transformer, attention-based
mechanisms 
by Young, et al.~\cite{You18}, bidirectional recurrent neural networks by Schuster
and Paliwal~\cite{Sch97}, temporal convolutional networks by Bai et al.(2018)~\cite{Bai18}, deep
Bayesian neural network, deep sequential spatio-temporal residual neural
network, dual self-attention network and memory-based
ordinal regression deep neural networks. 

The attention mechanism of any deep-learning framework has the ability to
select hidden states and patterns within a particular dataset which makes
them more attractive for modelling complex situations.  This mechanism may
even produce an unexpected performance when combined with unidirectional,
bidirectional or multidirectional mechanisms. The sequential robustness of
any framework is correlated with the nature of the problem set being resolved,
the selected dataset and the evaluation criteria. 

It is clear that the art of identification, selection, configuration, deployment
and evaluation of any deep-learning framework has not been exhausted. Furthermore
many of these frameworks have not been explained adequately and thus the
experimental work is unclear.

\subsubsection{Complex sequential datasets}
The majority of existing sequential data types listed  in Annexure 2 were
either univariate or multivariate or both and were sourced from different
domains such as: traffic, financial stock markets, meteorological weather 
and climate information, energy consumption, natural language sentiment 
processing, telecommunications, astronomy, etc.

\begin{figure}[!t]
\includegraphics[bb=20 300 800 590,scale=0.300]{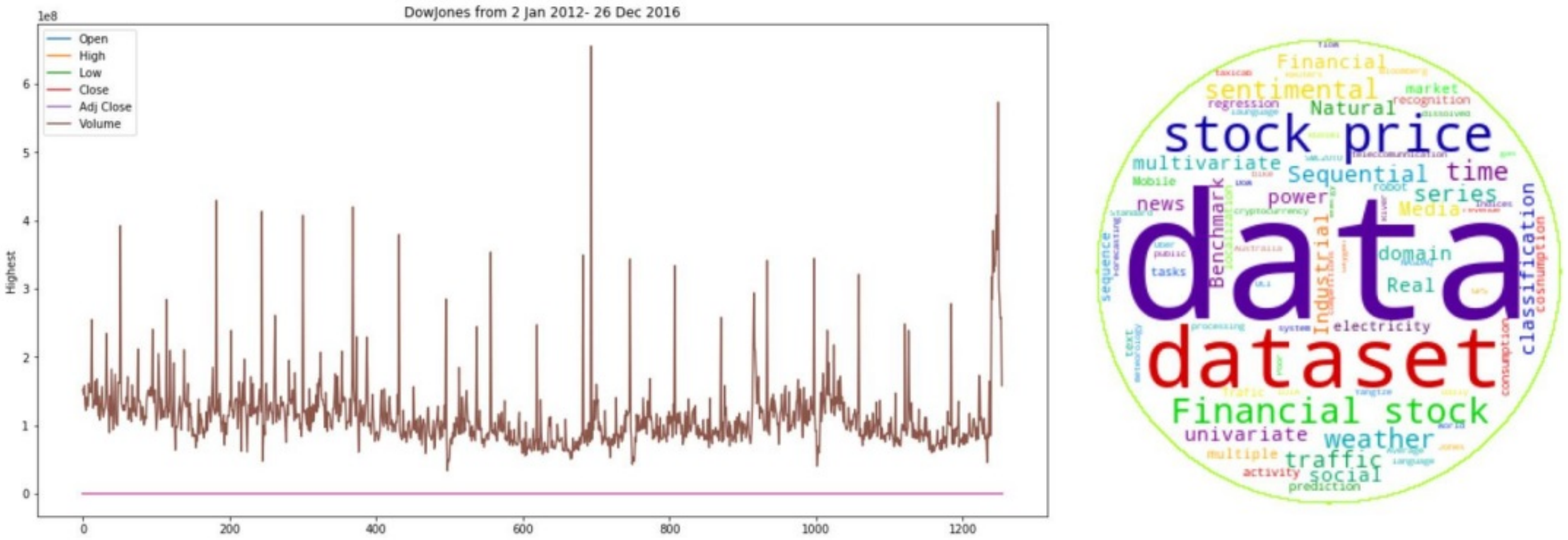}
\caption{Irregular patterns of the financial dataset}\label{figure:four}
\end{figure}

Figure~\ref{figure:four} shows experimental data analysis results showing irregular sequential
patterns of the financial dataset from NASDAQ stock market (2 January 2012--26
December 2016) by~\cite{Qia17} and the wordle analysis of sequential datasets.

\subsubsection{Evaluation}

Performance evaluation of most modern deep-learning frameworks is based on
one of the following mechanisms:  prediction accuracy, efficiency, correlation,
baseline, consistency, visualization sharpness, robustness and computational
complexity. There is no single framework that has been evaluated in terms
of its multidimensional performance in which critical issues such as efficiency,
accuracy, consistency, reliability, stability and transparency were considered. 

In terms of performance metrics, there are over 40 metrics listed in Annexure
3 which were applied in the evaluation processes of the different frameworks.
The systematic literature review process revealed that several of the accuracy
metrics were most often used.  The majority of these accuracy metrics
are categorised and measured based on absolute, squared, relative, symmetric
and percentage error types Bal et al.~\cite{Bal16}.  Mean absolute
error (MAE), mean square error (MSE) and root mean squared error (RMSE) dominated
this space since they are easy to compute~\cite{Ras20}. RMSE and MSE in
particular have sensitivity challenges associated with the stability of the
selected frameworks~\cite{Tan19}. Any results that yield the best
(lowest) RMSE value, demonstrate the stability of such an algorithm or framework
as they are less sensitive to outliers.

Selection of the correct performance evaluation metrics is central to the
identification of a proper framework for any given problem.  A particular
framework can be deemed sub-optimal because the wrong evaluation metrics
were chosen~\cite{Gun09}. When comparing the performance of state-of-the-art
baseline frameworks it may be necessary to consider applying different combinations
of evaluation criteria.

\subsection{Existing recommendations}

Future work recommendations from current studies point to the need
to improve on: the interpretability of modern deep-learning models, resolving
prediction problems associated with volatile time series data, and lowering
performance computing costs.  They also suggest the exploration of modern
deep-learning frameworks such as: adversarial networks, temporal convolution
networks, transformer networks, and CapsNets, by combining them with
other architectures such as the dilated CNNs. Furthermore model optimisation
should be explored through its manipulation of both internal and external
architectural properties, deploying future frameworks on complex univariate
and multivariate cases with exogenous environments.

\section{Discussion}

Existing work on deep-learning frameworks for sequential analysis has produced
a large pool of publications covering wide range of issues such as methods,
experimental design, optimisation techniques, input signal issues in the form
of datasets, application areas and theories and many others, however, there
are important systematic gaps, limitations and inconsistencies in these studies.
It is clear that researchers have different objectives when designing and
analysing deep-learning frameworks.  This creates an unnecessary discord
in identification of a systematic evaluation methodology. 

The deep-learning algorithms listed in Annexure 1 are suboptimal and not
efficient in analysing complex sequential environments.  They lack transparency,
interpretability, and their performance evaluation is not exhaustive.
These issues have not yet been adequately documented and publications contradict
one another. For example how can any experiment conclude that the designed
state-of-the-art framework architecture is the best when it only uses performance
accuracy based on one criterion?

Historical sequential datasets from financial stock markets have produced
understandable volatile, non-linear and chaotic characteristics.
The financial stock market domain is very sensitive
to changes such as the Covid-19 pandemic. It is highly correlated to these
financial events. This makes the domain an interesting environment for developing
an enhanced deep-learning framework for accurate analysis of sequential irregular
patterns. 

It is possible that a deep-learning framework that has been trained on 
complex datasets, when exposed to normal or ordinary environments, will 
outperform other frameworks in terms of robustness and efficiency. 

There is room for exploring better optimised sequential frameworks that have the
potential of producing a better sequential analysis performance. Current
deep-learning algorithms can be adapted to analyse irregular-patterned
complex sequential datasets as a means of improvement.  These new approaches 
need to focus on a more detailed cross cutting approach that address 
efficiency, accuracy, consistency and reliability issues.

\section{Annexures}

The annexures can be found on the GitHub website.
\url{https://github.com/Dandajena/SDA/blob/master/Annexures.pdf}

\section{Conclusion and future recommendations}
The goal of this research was to determine which deep-learning frameworks
are currently being used to analyse irregular-patterned complex
sequential datasets. Using a systematic literature research methodology the
issues associated with the performance of well-known state-of-the-art deep
learning frameworks for irregular-patterned complex sequential analysis and
their respective challenges were identified. 

It was found from existing literature that several researchers feel that
deep-learning algorithms are suboptimal and not efficient in analysing complex
sequential datasets~\cite{Pou18}. It is thus clear that there
is a need to improve the existing performance evaluation methods into a unified
multidimensional evaluation method.  It cannot be claimed that a state-of-the-art
framework is optimal without applying an extensive, transparent and traceable
performance evaluation procedure on the results of such a framework. 

To address this deficiency it is worth considering a combination of:
(1) selecting a proper algorithm architecture and redesigning it; 
(2) sensitively tuning it;
as well as (3) evaluating its performance in a multidimensional mode based
on complex irregular-patterned complex sequential datasets. 

This will provide a potential approach towards the development of a new
breed of robust deep-learning frameworks which are efficient, accurate, consistent,
reliable, stable and transparent. This can only be achieved by selecting
an existing sequential dataset on which research has already been done as
a way of minimising the experimental variables and parameters.  It will allow
comparison.

The systematic literature review results show that the financial stock market domain---particularly
from well-estab\-lished financial markets---provides irregular-patterned sequential
datasets.  Furthermore, it is important to rank their volatility before selecting
and adopting these datasets in complex sequential modelling experiments.
Improving the design of algorithms
based on such extreme complex sequential datasets provides much needed theoretical,
methodological and experimental contributions on the performance exploration
of deep-learning frameworks. Extreme scenarios are associated with big data
and it has engulfed our lives. 
Faster analysis 
through state-of-the-art frameworks may offer the much needed scientific, engineering,
academic and business solutions to make the world a better place~\cite{Sch16}.

Finally, to enrich a future experimental setup, it is advisable to consider
some best properties of the sequential recurrent neural networks proposed
by Zhang et al.~\cite{Zha185} as baseline architecture. This process could
be coupled with other novel architectures such as temporal convolution networks,
transformer networks, CapsNets, unidirectional as well as attention,
bidirectional or multidirectional mechanisms.

\begin{acks}
This work is supported by Research Committee of the University of the Western
Cape and the Telkom/Aria Technology Africa Centre of Excellence.
\end{acks}

\balance

\bibliographystyle{ACM-Reference-Format}
\bibliography{kudaSaicsit2020} 
\end{document}